\documentclass[runningheads]{llncs}
\usepackage{graphicx}
\usepackage[utf8x]{inputenc}
\usepackage{bbding}

\usepackage{xspace}
\makeatletter
\DeclareRobustCommand\onedot{\futurelet\@let@token\@onedot}
\def\@onedot{\ifx\@let@token.\else.\null\fi\xspace}

\def\etc{\emph{etc}\onedot}

\makeatother

\usepackage{comment}

\usepackage{amsmath}
\usepackage{amssymb}
   
\usepackage{multirow}

\newcommand{\our}{JiGen\xspace}

\usepackage[export]{adjustbox}

\usepackage{array}
\newcommand{\PreserveBackslash}[1]{\let\temp=\\#1\let\\=\temp}
\newcolumntype{C}[1]{>{\PreserveBackslash\centering}p{#1}}

\begin{document}
%
\title{Tackling Partial Domain Adaptation with Self-Supervision}

%
%

\author{Silvia Bucci\inst{1,2}\orcidID{0001-6318-7288}  \and 
Antonio D'Innocente\inst{2,3}\orcidID{0002-4902-9301} \and 
Tatiana Tommasi\inst{1}\orcidID{0001-8229-7159} \Envelope}
\authorrunning{S. Bucci et al.}
%
\institute{Politecnico di Torino, Italy\and
Istituto Italiano di Tecnologia, Italy\and
Sapienza Università di Roma, Italy  \\
\email{\{silvia.bucci, antonio.dinnocente\}@iit.it, tatiana.tommasi@polito.it}
}

\maketitle              
\begin{abstract}
Domain adaptation approaches have shown promising results in reducing the marginal distribution difference
among visual domains. They allow to train reliable models that work over datasets of different nature (photos, paintings \etc), but they still struggle when the domains do not share an identical label space. 
In the partial domain adaptation setting, where the target covers only a subset of the source classes, it is 
challenging to reduce the domain gap without incurring in negative transfer. Many solutions just keep the standard 
domain adaptation techniques by adding heuristic sample weighting strategies. In this work we show how the 
self-supervisory signal obtained from the spatial co-location of patches can be used to define a side task 
that supports adaptation regardless of the exact label sharing condition across domains. 
We build over a recent work that introduced a jigsaw puzzle task for domain generalization: we
describe how to reformulate this approach for partial domain adaptation and we show how it boosts
existing adaptive solutions when combined with them. The obtained experimental results on 
three datasets supports the effectiveness of our approach.
\end{abstract}
\section{Introduction}

Today the most popular synonym of \emph{Artificial Intelligence} is \emph{Deep Learning}: 
new convolutional neural network architectures constantly hit the headlines by improving the state of the 
art for a wide variety of machine learning problems and applications with impressive results.
The large availability of annotated data, as well as the assumption of training and testing
on the same domain and label set, are important ingredients of this success. However
this closed set condition is not realistic and the learned models cannot be said fully \emph{intelligent}. 
Indeed, when trying to summarize several definitions of intelligence from dictionaries, psychologists 
and computer scientists of the last fifty years, it turns out that all of them highlight as fundamental 
the  ability to adapt and achieve goals in a wide range of environments and conditions \cite{AI}.
\emph{Domain Adaptation} (DA) and \emph{Domain Generalization} (DG) methods are trying to go over this 
issue and allow the application of deep learning models in the wild. Many DA and DG approaches have been 
developed for the object classification task to reduce the domain gap across samples obtained from different
acquisition systems, different illumination conditions and visual styles, but most of them keep a strong 
control on the class set, supposing that the trained model will be deployed exactly on the same categories 
observed during training. When part of the source classes are missing at test time, those models show 
a drop in performance which indicates the effect of negative transfer in this 
\emph{Partial Domain Adaptation} (PDA) setting. 
The culprit must be searched in the need of solving two challenging tasks at the same time: one that exploits
all the available source labeled data to train a reliable classification model in the source domain 
and another that estimates and minimizes the marginal distribution difference between source and target, 
but disregards the potential presence of a conditional distribution shift. 
Very recently it has been shown that this second task may be substituted with self-supervised objectives 
which are agnostic with respect to the domain identity of each sample. 
In particular, \cite{jigen} exploits image patch shuffling 
and reordering as a side task over multiple sources: it leverages the intrinsic regularity 
of the spatial co-location of patches and generalizes to new domains. This information appears 
also independent from the specific class label of each image, which makes it an interesting 
reference knowledge also when the class set of source and target are only partially overlapping. 
We dedicate this work to investigate how 
the jigsaw puzzle task of \cite{jigen} performs in the PDA setting and how it can be
reformulated to reduce the number of needed learning parameters. 
The results on three different datasets indicate that our approach outperforms several competitors 
whose adaptive solutions include specific strategies to 
down-weight the samples belonging to classes supposedly absent from the target. We also discuss 
how such a re-scaling  process can be combined with the jigsaw puzzle obtaining further 
gains in performance.

\section{Related Work}

\noindent
\textbf{Closed Set Domain Adaptation}
When the source and target data belongs to two different marginal distributions but the two domains share the same label set,  
it is relatively easy to train a source classifier that adapts to the target domain by adding extra conditions on the learned features. Several recent approaches minimize domain shift measures like the Maximum Mean Discrepancy \cite{Tzeng:MMD:arxiv14,Long_icml15,long2016unsupervised,LongZ0J17}, and the Wasserstein distance \cite{deepJDOT,slicedWasserstein_cvpr19}, or exploit other statistical moment matching constraints \cite{Zellinger_CMD17,morerio2018minimalentropy} or even introduce dedicated batch normalization layers in deep learning networks \cite{carlucci2017auto,mancini2018boosting}.
Another family of methods use adversarial losses that force the data to be indistinguishable in terms of their domain label \cite{Ganin:DANN:JMLR16,Hoffman:Adda:CVPR17}. Those solutions borrow the idea at the basis of Generative Adversarial Network (GAN, \cite{Goodfellow:GAN:NIPS2014}) that can be also directly applied to match domains at pixel level \cite{Bousmalis:Google:CVPR17,sankaranarayanan2017generate,russo17sbadagan}.
All these methods exploit the availability of unsupervised target data at training time by leveraging on the domain identity of the samples. However, several other unsupervised models could be learned from those samples and used as extra regularization tools for the source model. A very common solution is that of measuring the source prediction uncertainty on the target
data with an entropy loss which is minimized during training \cite{long2016unsupervised,luo2017label}. 
A recent stream of works has introduced techniques to extract self-supervisory signals from unlabeled data as the patch relative position \cite{DoerschGE15,NorooziF16}, counting primitives \cite{learningtocount}, or image coloring \cite{zhang2016colorful}. They capture invariances and regularities that allow to train models useful as
fine-tuning priors, and those information appear also independent from the specific visual domain of the data 
from which they are obtained. Indeed, \cite{jigen} showed how shuffling and reordering image patches can be used as a side task to learn a robust model over 
multiple sources that generalizes even to unseen target samples.

\noindent 
\textbf{Partial Domain Adaptation}
The PDA setting relaxes the fully shared label space assumption among the domains and allows the target to cover only a subset of the source class set. Here it becomes important to adjust the adaptation process so that the samples with not shared labels would not influence the learning process.
The first work which considered this setting focused on \emph{localizing domain specific and generic image regions} \cite{LOAD_ICRA}. The attention maps produced by this initial procedure are less sensitive to the difference in class set with respect to the standard domain classification procedure and
allow to guide the training of a robust source classification model.
Although suitable for robotics applications, this solution is insufficient when each domain has spatially diffused characteristics. 
In those cases the more commonly used PDA technique consists in adding a \emph{re-weight source sample strategy} to a standard domain adaptation learning process. Both the Selective Adversarial Network (SAN, \cite{SAN}) and the Partial Adversarial Domain Adaptation (PADA, \cite{PADA_eccv18}) approaches build over the domain-adversarial neural network architecture \cite{Ganin:DANN:JMLR16} and exploit the source classification 
model predictions on the target samples to evaluate a statistics on the class distribution. The estimated contribution of each source class  either weights the class-specific domain classifiers \cite{SAN}, or re-scales the respective classification loss and a single overall domain classifier \cite{PADA_eccv18}. A different solution is proposed in \cite{IWAN}, where each domain has its own feature
extractor and the source sample weight is obtained from the domain recognition model rather than from the source classifier.
An alternative view on the PDA problem is presented in two recent preprints \cite{TWIN_PDA,featurenorm_PDA}. The first work uses two separate deep classifiers to reduce the domain shift by enforcing a minimal inconsistency between their predictions on the target. Moreover the class-importance weight is formulated analogously to PADA, but averaging over the output of both the source classifiers. The second work does not attempt to aligning the whole domain distributions and focuses instead on matching the feature norm of source and target. This choice makes the proposed approach robust to negative transfer with good results in the PDA setting without any heuristic weighting mechanism.

Our work follow this research direction seeking a different solution with respect to the usual adversarial and sample
weighting technique. We propose to leverage the self-supervised signal captured by a jigsaw puzzle task on the image patches as side objective to the classification model and show its effectiveness both alone and in combination
with other more standard strategies.

\section{Solving Jigsaw Puzzles for Partial Domain Adaptation}

\begin{figure}[t]
  \centering
\includegraphics[width=1.07\textwidth]{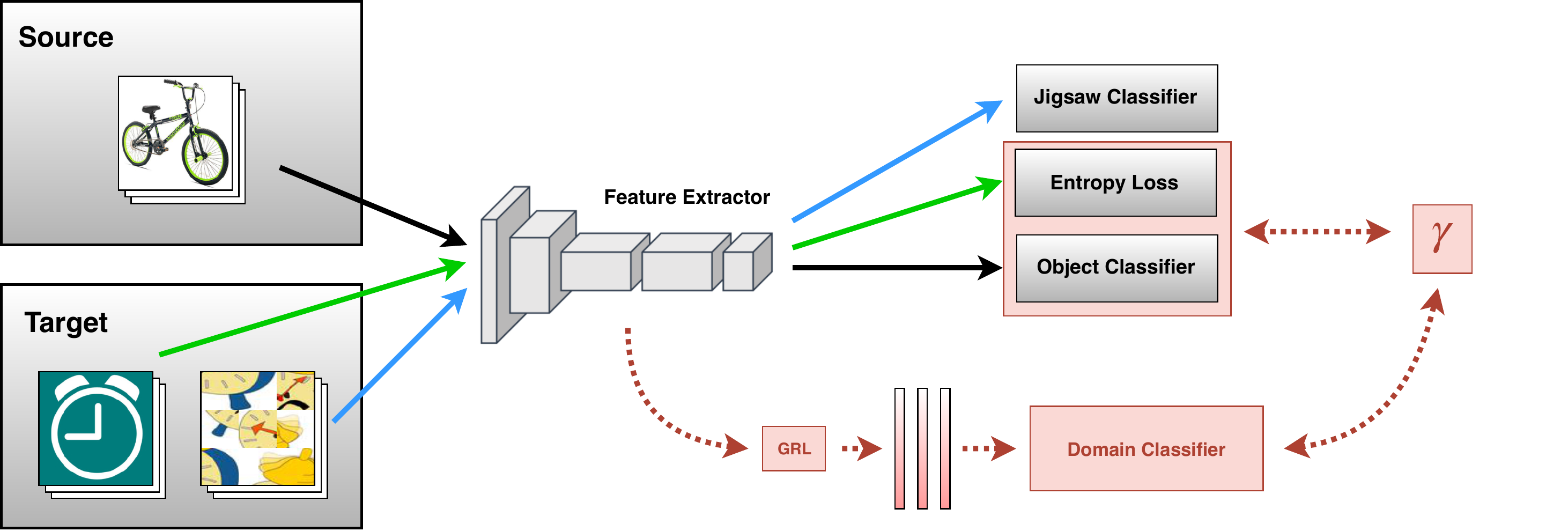}
\caption{Schematic representation of our SSPDA approach. All the parts in gray describe the main blocks 
of the network with the solid line arrows indicating the contribution of each group of training samples 
to the corresponding final tasks and related optimization objectives according to the assigned blue/green/black
colors. 
The blocks in red illustrate the domain adversarial 
classifier with the gradient reversal layer (GRL) and source sample weighting procedure (weight $\gamma$) that can be added to SSPDA (refer to section \ref{otherpda}).}
\label{fig:scheme}\vspace{-2mm}
\end{figure}

\subsection{Problem Setting}
Let us introduce the technical terminology for the PDA scenario. We have $n^s$ annotated samples from a source domain $\mathcal{D}_s = \{(\mathbf{x}_i^s,\mathbf{y}_i^s)\}_{i=1}^{n^s}$, drawn from the distribution $S$, and $n^t$ unlabeled examples of the target domain $\mathcal{D}_t = \{\mathbf{x}_j^t\}_{j=1}^{n^t}$ drawn from a different distribution $T$. The label space of the target domain is contained in that of the source domain $\mathcal{Y}_t \subseteq \mathcal{Y}_s$. Thus, besides dealing with the marginal shift $S\neq T$ as in standard unsupervised domain adaptation, it is necessary to take care of the difference in the label space which makes the problem even more challenging. If this information is neglected and the matching between the whole source and target data is forced, any adaptive method may incur in a degenerate case producing worse performance than its plain non-adaptive version. Still the objective remains that of learning both class discriminative and domain invariant feature models 
which can be formulated as a multi-task learning problem \cite{Caruana:1997}. Instead of just focusing on the explicit reduction of the feature domain discrepancy, one could consider some inherent characteristics shared by any visual domain regardless of the assigned label and derive a learning problem to solve together with the main classification task. By leveraging the inductive bias of related objectives, multi-task learning regularizes the overall model and improves generalization having as an
implicit consequence the reduction of the domain bias. This reasoning is at the basis of the recent work \cite{jigen}, which proposed to use jigsaw puzzle as a side task for closed set domain adaptation and generalization: the model named JiGen is described in details in the next subsection.

\subsection{Jigsaw Puzzle Closed Set Adaptation}
Starting from the $n^s$ labeled and $n^t$ unlabeled images, the method in \cite{jigen} decomposes them according to an $3\times 3$ grid obtaining $9$ squared patches from every sample, which are then moved from their original location and re-positioned randomly to form a shuffled version $\mathbf{z}$ of the original image $\mathbf{x}$. Out of all the $9!$ possibilities, a set of $p=1,\ldots, P$ permutations are chosen on the basis of their maximal reciprocal Hamming distance \cite{NorooziF16} and used to define a jigsaw puzzle classification task which consists in recognizing the index $p$ of the permutation used to scramble a certain sample.
All the original $\{(\mathbf{x}_i^s,\mathbf{y}_i^s)\}_{i=1}^{n^s}$, $\{\mathbf{x}_j^t\}_{j=1}^{n^t}$ as well as the shuffled versions of the images $\{(\mathbf{z}_k^s,\mathbf{p}_k^s)\}_{k=1}^{K^s}$ , $\{(\mathbf{z}_k^t,\mathbf{p}_k^t)\}_{k=1}^{K^t}$ 
are given as input to a multi-task
deep network where the convolutional feature extraction backbone is indicated by $G_f$ and is parametrized by $\theta_f$, 
while the classifier $G_c$ of the object labels and $G_p$ of the permutation indices, are parametrized respectively by $\theta_c$ and $\theta_p$. The source samples are involved both in the object classification and in the jigsaw puzzle classification task, while the unlabeled target samples deal only with the puzzle task. To further exploit the available target data, the uncertainty of the estimated prediction $\hat{\mathbf{y}}^t=G_c(G_f(\mathbf{x}^t))$ is evaluated through
the entropy $H  
=-\sum_{l=1}^{|\mathcal{Y}_s|}\hat{y}^t_{l} \log \hat{y}^t_{l}$
and minimized to enforce the decision boundary to pass through low-density areas.
Overall the end-to-end JiGen multi-task network is trained by optimizing the following objective 
\begin{align} \small
    \arg \min_{\theta_f, \theta_c, \theta_p} ~~ & 
    \frac{1}{n^s} \sum_{i=1}^{n^s}   \mathcal{L}_c(G_c(G_f(\mathbf{x}^s_i),y_i^s)) +  \alpha_s \frac{1}{K^s}\sum_{k=1}^{K^s}\mathcal{L}_p(G_p(G_f(\mathbf{z}^s_k),p_k^s)) +  \nonumber \\ 
      &
    \eta~\frac{1}{n^t} \sum_{j=1}^{n^t}  H(G_c(G_f(\mathbf{x}^t_j))) +  \alpha_t \frac{1}{K^t}
    \sum_{k=1}^{K^t}\mathcal{L}_p(G_p(G_f(\mathbf{z}^t_k),p_k^t)) ~,
    \label{equationJIGEN}
\end{align}
where $\mathcal{L}_c$ and $\mathcal{L}_p$ are cross entropy losses for both the object and puzzle classifiers. In the closed set scenario, the experimental evaluation of \cite{jigen} 
showed that tuning two different hyperparameters $\alpha_s$ and $\alpha_t$ respectively for the source and target puzzle classification loss is beneficial with respect to just using a single value $\alpha=\alpha_s=\alpha_t$, while it is enough to assign a small value to $\eta$ ($\sim 10^{-1}$). 

\subsection{Jigsaw Puzzle for Partial Domain Adaptation}
The two $\mathcal{L}_p$ terms in (\ref{equationJIGEN}) provide a domain shift reduction effect on the learned feature representation, however their co-presence seem redundant: indeed the features are already chosen to minimize the 
source classification loss and the self-supervised jigsaw puzzle task on the target back-propagates its effect 
directly on the learned features inducing a cross-domain adjustment. By following this logic, we decided to drop the source jigsaw puzzle term, which corresponds to setting $\alpha_s=0$. This choice has a double positive effect: on one side it allows to reduce the number of hyper-parameters in the learning process leaving space for the introduction of other complementary learning conditions, on the other we let the self-supervised module focus only on the samples from the target without involving the extra classes of the source. In the following we indicate this approach as SSPDA: \emph{Self-Supervised Partial Domain Adaptation}. A schematic illustration of the method is presented in Figure \ref{fig:scheme}.

\subsection{Combining Self-Supervision with other PDA Strategies}
\label{otherpda}
To further enforce the focus on the shared classes, SSPDA can be extended to integrate a weighting mechanism analogous to that presented in \cite{PADA_eccv18}. The source classification output on the target data are accumulated as follow 
$\boldsymbol{\gamma} = \frac{1}{n^t}\sum_{j=1}^{n^t}\hat{\mathbf{y}}_j^t$
and normalized $\boldsymbol{\gamma} \leftarrow \boldsymbol{\gamma}/ \max(\boldsymbol{\gamma})$, obtaining a $|\mathcal{Y}_t|$-dimensional vector that quantifies the contribution of each source class. 
Moreover, we can easily integrate a domain discriminator $G_d$ with a gradient reversal layers as in \cite{Ganin:DANN:JMLR16}, and adversarially maximize the related binary cross-entropy to increase the domain confusion, taking also into consideration the defined class weighting procedure for the source samples. In more formal terms, the final objective of our multi-task problem is\vspace{-2mm}
\begin{align} \small
    \arg \min_{\theta_f, \theta_c, \theta_p} \max_{\theta_d} ~~ 
    \frac{1}{n^s} \sum_{i=1}^{n^s}  \gamma_{y} \Big( \mathcal{L}_c(G_c(G_f(\mathbf{x}^s_i),y_i^s)) +  \lambda \log (G_d(G_f(\mathbf{x}^s_i))) \Big)+  \nonumber \\ 
    \frac{1}{n^t} \sum_{j=1}^{n^t} \gamma_{y} \Big( \eta  H(G_c(G_f(\mathbf{x}^t_j))) +  \lambda \log (1 - G_d(G_f(\mathbf{x}^t_j)))\Big)+ \nonumber \\
    \alpha_t \frac{1}{K^t}
    \sum_{k=1}^{K^t}\mathcal{L}_p(G_p(G_f(\mathbf{z}^t_k),p_k^t))~,
    \label{equationOUR}
\end{align}

\noindent where 
$\lambda$ is a hyper-parameter that adjusts the importance of the introduced domain discriminator. We adopted the same scheduling of \cite{Ganin:DANN:JMLR16} to update the value of $\lambda$, so that the importance of the domain discriminator increases with the training epochs,
avoiding the noisy signal at the early stages of the learning procedure. 
When $\lambda=0$ and $\gamma_{y} = 1/|\mathcal{Y}_s| $ we fall back to SSPDA. 

\section {Experiments}

\subsection{Datasets}
We test our algorithm on three different Partial Domain Adaptation benchmarks following the setting previously used in \cite{PADA_eccv18}.

\vspace{1mm}
\textbf{Office-31} \cite{Saenko:2010} is widely used in domain adaptation, it contains 4.652 images of 31 object categories common in office environments. Samples are drawn from three annotated distributions: Amazon (A), Webcam (W) and DSLR (D): we considered six different conditions by alternatively selecting one source domain and one target domain from AWD, and testing only 10 categories of the target which are those shared by Office-31 and Caltech-256. 

\vspace{1mm}
\textbf{Office-Home} \cite{venkateswara2017Deep} is a domain adaptation dataset containing around 15,500 images organized in 65 categories of common home and office objects. It has four domains: Art (Ar), Clipart (Cl), Product(Pr) and Real world (Rw), and is more challenging compared to Office-31 due to strong domain shifts in distributions, class imbalances within the data and size variations of images. We  considered 12 different settings by choosing source and target domain from the available domains, and removed from the target the last 40 classes in alphabetic order.

\vspace{1mm}
\textbf{VisDA2017} is the dataset used in the 2017 Visual Domain Adaptation challenge (classification track). It has two domains, synthetic 2D object renderings and real images with a total of 208k images organized in 12 categories. In our experiments we focused on the synthetic-to-real shift, the same considered in the original challenge, but keeping
only the first 6 categories of the target in alphabetic order.
With respect to the other considered testbeds, VisDA2017 allow us to investigate our approach on a very large-scale sample size scenario.

\subsection{Implementation Details}
We implemented all our deep methods in PyTorch. Specifically the main backbone of our SSPDA network is a ResNet-50 pre-trained on ImageNet and corresponds to the feature extractor defined as $G_f$, while the specific object and puzzle classifiers $G_c,G_p$ are implemented each by an ending fully connected layer. The domain classifier $G_d$ is introduced by adding three fully connected layers after the last pooling layer of the main backbone, and using a sigmoid function for the last activation as in \cite{Ganin:DANN:JMLR16}. By training the network end-to-end we fine-tune all the feature layers, while $G_c, G_p$ and $G_d$ are learned from scratch. We train the model with backpropagation using SGD with momentum set at $0.9$, weight decay  $0.0005$ and initial learning rate $0.0005$. We use a batch size of 64 (32 source samples + 32 target samples) and, following \cite{jigen}, we shuffle the tiles of each input image with probability $1 - \beta$, with $\beta = 0.7$. Shuffled samples are only used for the auxiliary jigsaw task, therefore only unshuffled (original) samples are passed to $G_d$ and $G_c$ for domain and label predictions. The entropy weight $\eta$ and jigsaw task weight $\alpha_t$ are set respectively to 0.2 and 1. Our data augmentation protocol is the same of \cite{jigen}. 

\vspace{2mm}\noindent
\textbf{Model Selection}  
As standard practice, we used $10\%$ of the source training domain to define a validation set on which the model is evaluated after each epoch $e$. The obtained accuracy $A_e$ is dynamically averaged with the value obtained at the previous epoch with $A_e\leftarrow w A_{e-1} + (1-w) A_{e}$. The final model to apply on the target is chosen as
the one producing the top accuracy over all the epochs $e=1,\ldots, E$. We noticed that this procedure leads to a more reliable selection of the best trained model, preventing to choose one that might have overfitted on the validation set. For all our experiments we kept $w=0.6$.
We underline that this smoothing procedure was applied uniformly on all our experiments. Moreover the 
hyper-parameters of our model are the same for \textbf{all} the domain pairs within 
each dataset and also across all the datasets. In other words we did \textbf{not} select a tailored set of parameters for
each sub-task of a certain dataset which could lead to further performance gains, a procedure used 
in previous works \cite{PADA_eccv18,SAN}.

\subsection{ Results of SSPDA}
Here we present and discuss the obtained classification accuracy results on the three considered datasets: Office-31 in Table \ref{tab:office31}, Office-Home in Table \ref{tab:office-home} and VisDA in Table \ref{tab:visda2017}. Each table is
organized in three horizontal blocks: the first one shows the results obtained with standard DA methods, the second block illustrates the performance with algorithms designed to deal with PDA and the third one includes the scores of \our and SSPDA. Only Table \ref{tab:office31} has an extra fourth block that we will discuss in details in the following section.

Both \our and SSPDA exceed all plain DA methods and present accuracy value comparable to those of the PDA methods. In particular
SSPDA is always better than PADA \cite{PADA_eccv18} on average, and for both Office-Home and VisDA it also outperforms all the other competing PDA methods with the only exception of IAFN \cite{featurenorm_PDA}. We highlight that this approach uses a competitive version of ResNet-50 as backbone, with extra bottleneck fully connected layers which add about 2 million 
parameters to the standard version of ResNet-50 that we adopted.

\begin{table}[t]
\centering
\caption{Classification accuracy in the PDA setting defined on the Office-31 dataset with all the 31 classes used for each source domain, and a fixed set of 10 classes used for each target domain. The results are obtained using 10 random crop predictions on each target image and are averaged over three repetitions of each run.}
\begin{tabular}{l@{~~~}C{1.3cm}C{1.3cm}C{1.3cm}C{1.3cm}C{1.3cm}C{1.3cm}|C{1.3cm}}
\hline
       &       \multicolumn{6}{c|}{\textbf{Office-31}} &       \\
       & \textbf{A}$\rightarrow$\textbf{W} & \textbf{D}$\rightarrow$\textbf{W} & \textbf{W}$\rightarrow$\textbf{D} & \textbf{A} $\rightarrow$\textbf{D} & \textbf{D}$\rightarrow$\textbf{A} & \textbf{W}$\rightarrow$\textbf{A} & \textbf{Avg.} \\
\hline
Resnet-50           & $75.37$ & $94.13$ & $98.84$ & $79.19$ & $81.28$ & $85.49$ & $85.73$ \\ 
\hline
DAN\cite{Long:2015}               & $59.32$ & $73.90$ & $90.45$ & $61.78$ & $74.95$ & $67.64$ & $71.34$ \\ 
DANN\cite{Ganin:DANN:JMLR16}      & $75.56$ & $96.27$ & $98.73$ & $81.53$ & $82.78$ & $86.12$ & $86.50$ \\ 
ADDA\cite{Hoffman:Adda:CVPR17}    & $75.67$ & $95.38$ & $99.85$ & $83.41$ & $83.62$ & $84.25$ & $87.03$ \\ 
RTN\cite{long2016unsupervised}    & $78.98$ & $93.22$ & $85.35$ & $77.07$ & $89.25$ & $89.46$ & $85.56$ \\ 
\hline
IWAN \cite{IWAN}                & $89.15$ & $99.32$ & $99.36$ & $90.45$ & $\textbf{95.62}$ & $94.26$ & $94.69$ \\ 
SAN \cite{SAN}                  & $93.90$ & $99.32$ & $99.36$ & $94.27$ & $94.15$ & $88.73$ & $94.96$ \\ 
PADA\cite{PADA_eccv18}            & $86.54$ & $\mathbf{99.32}$ & $\mathbf{100}$ & $82.17$ & $92.69$ & $\mathbf{95.41}$ & $92.69$ \\ 
TWIN \cite{TWIN_PDA}    & $86.00$ & $99.30$ & $\mathbf{100}$ & $86.80$ & $94.70$ & $94.50$ & $93.60$ \\

\hline\hline
\our \cite{jigen} & $ 92.88 $ & $92.43$   & $98.94$  & $89.6$  & $84.06$  & $92.94$    &   $91.81$  \\
SSPDA & $91.52$ & $92.88$  & $98.94$  & $90.87$  & $90.61$  & $94.36$   &  $93.20$  \\
\hline
SSPDA-$\gamma$ & $99.32$ & $94.69$  & $99.36$  & $96.39$  & $86.36$  & $94.22$   & $95.06$    \\
SSPDA-PADA & $\textbf{99.66}$ & $94.46$  & $99.57$  &   $\textbf{97.67}$ & $87.33$  & $94.26$   &  $\textbf{95.49}$   \\
\hline
\end{tabular}
\vspace{-3mm}
    \label{tab:office31} 
\end{table}
\begin{table}[t]
\caption{Classification accuracy in the PDA setting defined on the Office-Home dataset with all the 65 classes used for each source domain, and a fixed set of 25 classes used for each target domain.
The results are obtained by averaging over three repetitions of each run.}
\centering
\begin{adjustbox}{width=1\textwidth}
\begin{tabular}{l@{~~~}C{1.4cm}C{1.4cm}C{1.4cm}C{1.4cm}C{1.4cm}C{1.4cm}C{1.4cm}C{1.4cm}C{1.4cm}C{1.4cm}C{1.4cm}C{1.4cm}|C{1.4cm}}
\hline
       &       \multicolumn{12}{c|}{\textbf{Office-Home}} &       \\
       & \textbf{Ar}$\rightarrow$\textbf{Cl} & \textbf{Ar}$\rightarrow$\textbf{Pr} & \textbf{Ar}$\rightarrow$\textbf{Rw} & \textbf{Cl} $\rightarrow$\textbf{Ar} & \textbf{Cl}$\rightarrow$\textbf{Pr} & \textbf{Cl}$\rightarrow$\textbf{Rw} & \textbf{Pr}$\rightarrow$\textbf{Ar} & \textbf{Pr}$\rightarrow$\textbf{Cl}  & \textbf{Pr}$\rightarrow$\textbf{Rw}  & \textbf{Rw}$\rightarrow$\textbf{Ar}  & \textbf{Rw}$\rightarrow$\textbf{Cl}  & \textbf{Rw}$\rightarrow$\textbf{Pr} &  \textbf{Avg.} \\
\hline
Resnet-50          & $38.57$ & $60.78$ & $75.21$ & $39.94$ & $48.12$ & $52.90$ & $49.68$ & $30.91$ & $70.79$ & $65.38$ & $41.79$ & $70.42$ & $53.71$ \\ 
\hline
DAN\cite{Long:2015}               & $44.36$ & $61.79$ & $74.49$ & $41.78$ & $45.21$ & $54.11$ & $46.92$ & $38.14$ & $68.42$ & $64.37$ & $45.37$ & $68.85$ & $54.48$ \\
DANN\cite{Ganin:DANN:JMLR16}      & $44.89$ & $54.06$ & $68.97$ & $36.27$ & $34.34$ & $45.22$ & $44.08$ & $38.03$ & $68.69$ & $52.98$ & $34.68$ & $46.50$ & $47.39$ \\
RTN\cite{long2016unsupervised}   & $49.37$ & $64.33$ & $76.19$ & $47.56$ & $51.74$ & $57.67$ & $50.38$ & $41.45$ & $75.53$ & $70.17$ & $51.82$ & $74.78$ & $59.25$ \\
\hline
IWAN \cite{IWAN}                 & $53.94$ & $54.45$ & $78.12$ & $61.31$ & $47.95$ & $63.32$ & $54.17$ & $52.02$ & $81.28$ & $\textbf{76.46}$ & $56.75$ & $\textbf{82.90}$ & $63.56$\\ 
SAN \cite{SAN}                   & $44.42$ & $68.68$ & $74.60$ & $67.49$ & $64.99$ & $\textbf{77.80}$ & $59.78$ & $44.72$ & $80.07$ & $72.18$ & $50.21$ & $78.66$ & $65.30$\\ 
PADA\cite{PADA_eccv18}           & $51.95$ & $67.00$ & $78.74$ & $52.16$ & $53.78$ & $59.03$ & $52.61$ & $43.22$ & $78.79$ & $73.73$ & $56.60$ & $77.09$ & $62.06$\\
HAFN\cite{featurenorm_PDA}           & $53.35$ & $72.66$ & $80.84$ & $64.16$ & $65.34$ & $71.07$ & $66.08$ & $51.64$ & $78.26$ & $72.45$ & $55.28$ & $79.02$ & $67.51$\\
IAFN\cite{featurenorm_PDA}           & $\textbf{58.93}$ & $\textbf{76.25}$ & $\textbf{81.42}$ & $\textbf{70.43}$ & $\textbf{72.97}$ & $77.78$ & $\textbf{72.36}$ & $\textbf{55.34}$ & $80.40$ & $75.81$ & $60.42$ & $79.92$ & $\mathbf{71.83}$\\
\hline\hline
\our \cite{jigen} & $53.19$ & $65.45$ & $81.30$ & $68.84$ & $58.95$ & $74.34$ & $69.94$ & $50.95$ & $\textbf{85.38}$ & $75.60$ & $60.02$ & $81.96$ & $68.83$    \\
SSPDA & $52.02$ & $63.64$ & $77.95$ & $65.66$ & $59.31$ & $73.48$ & $70.49$ & $51.54$ & $84.89$ & $76.25$ & $\textbf{60.74}$ & $80.86$ & $68.07$ \\
\hline
\end{tabular}
\end{adjustbox}
    \label{tab:office-home} 
\end{table}

\begin{table}[t]
    \caption{Classification accuracy in the PDA setting defined on VisDA2017 dataset with all the 12 classes used for each source domain, and a fixed set of 6 classes used for each target domain.
    The results are obtained using 10 random crop predictions on each target image and are averaged over three repetitions of each run.}
\centering
\begin{tabular}{l@{~~~}C{3cm}} 
\hline
     \multicolumn{2}{c}{\textbf{VisDA2017}}        \\
       & \textbf{Syn.}$\rightarrow$\textbf{Real} \\
\hline
Resnet-50          & $45.26$ \\ 
\hline
DAN\cite{Long:2015}               & $47.60$ \\
DANN\cite{Ganin:DANN:JMLR16}      & $51.01$  \\
RTN\cite{long2016unsupervised}   & $50.04$ \\
\hline
PADA\cite{PADA_eccv18}           & $53.53$ \\
HAFN\cite{featurenorm_PDA} & $65.06$ \\
IAFN\cite{featurenorm_PDA} & $67.65$\\
\hline\hline
\our \cite{jigen} & $68.33$ \\
SSPDA & $\textbf{68.89}$ \\
\hline
\end{tabular}
\label{tab:visda2017} 
\end{table}


\begin{figure}[h!]
\centering
\begin{tabular}{c@{~~~}c@{~~~}c}
     \includegraphics[width=0.3\textwidth]{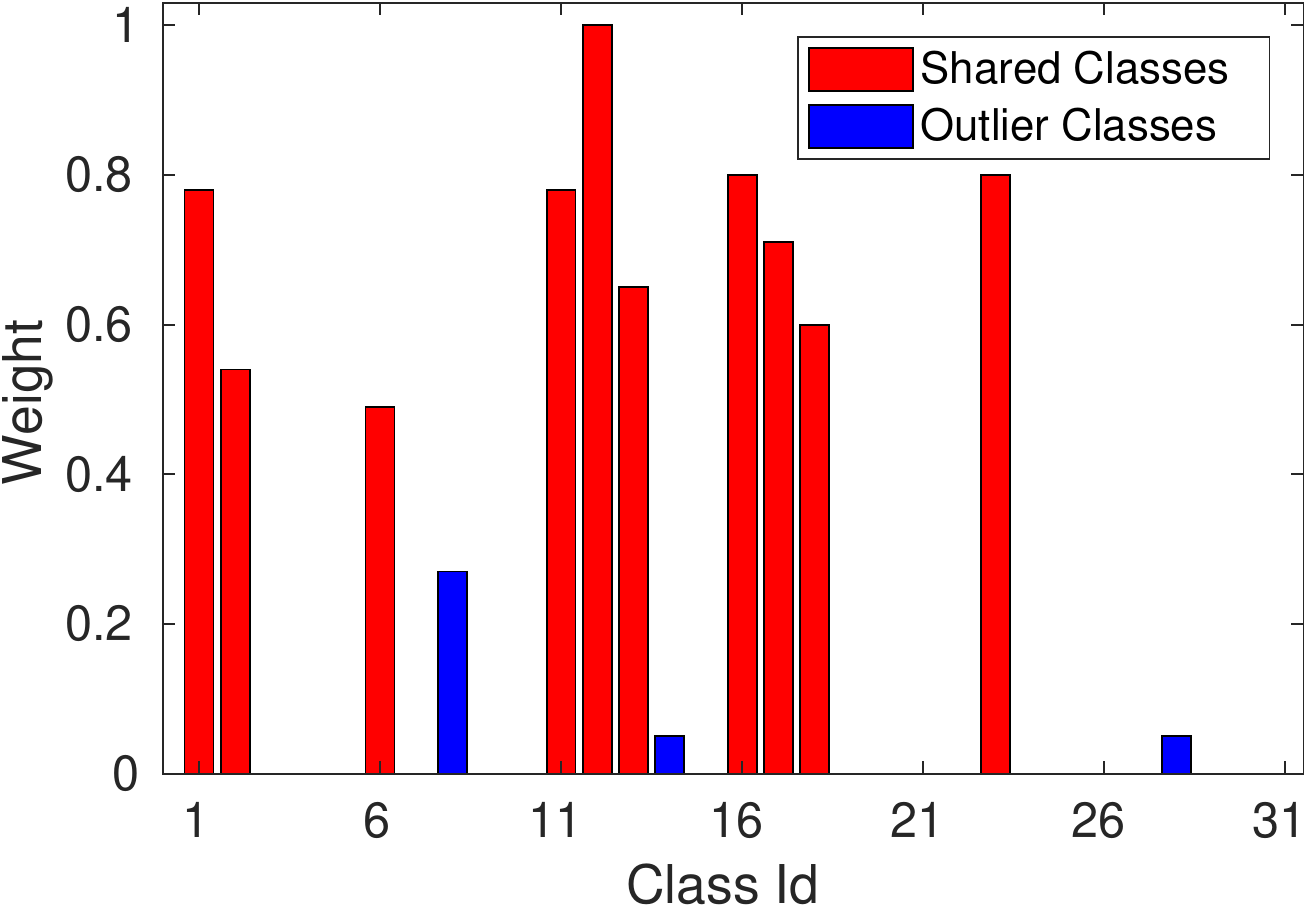} &  \includegraphics[width=0.3\textwidth]{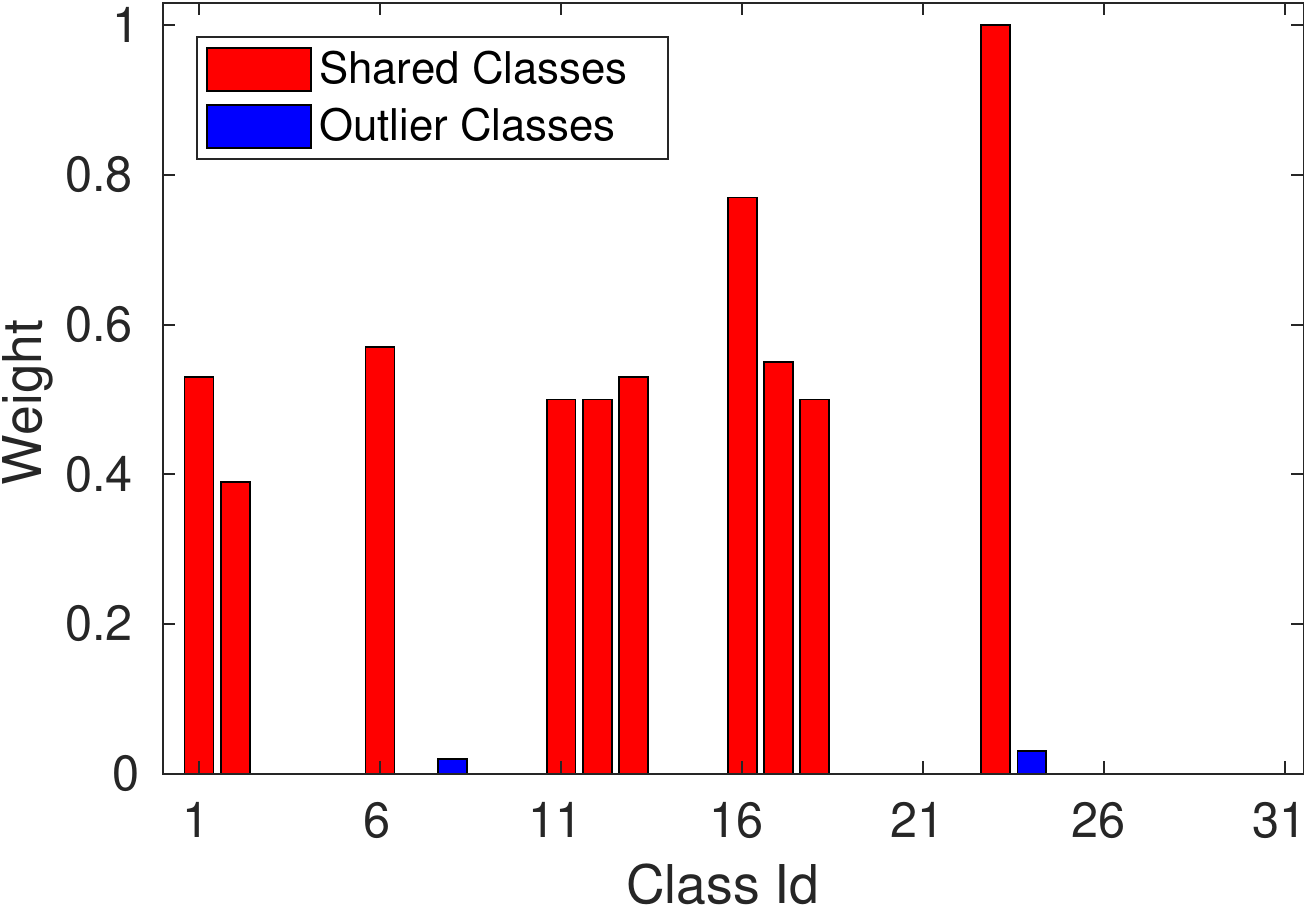} &
     \includegraphics[width=0.3\textwidth]{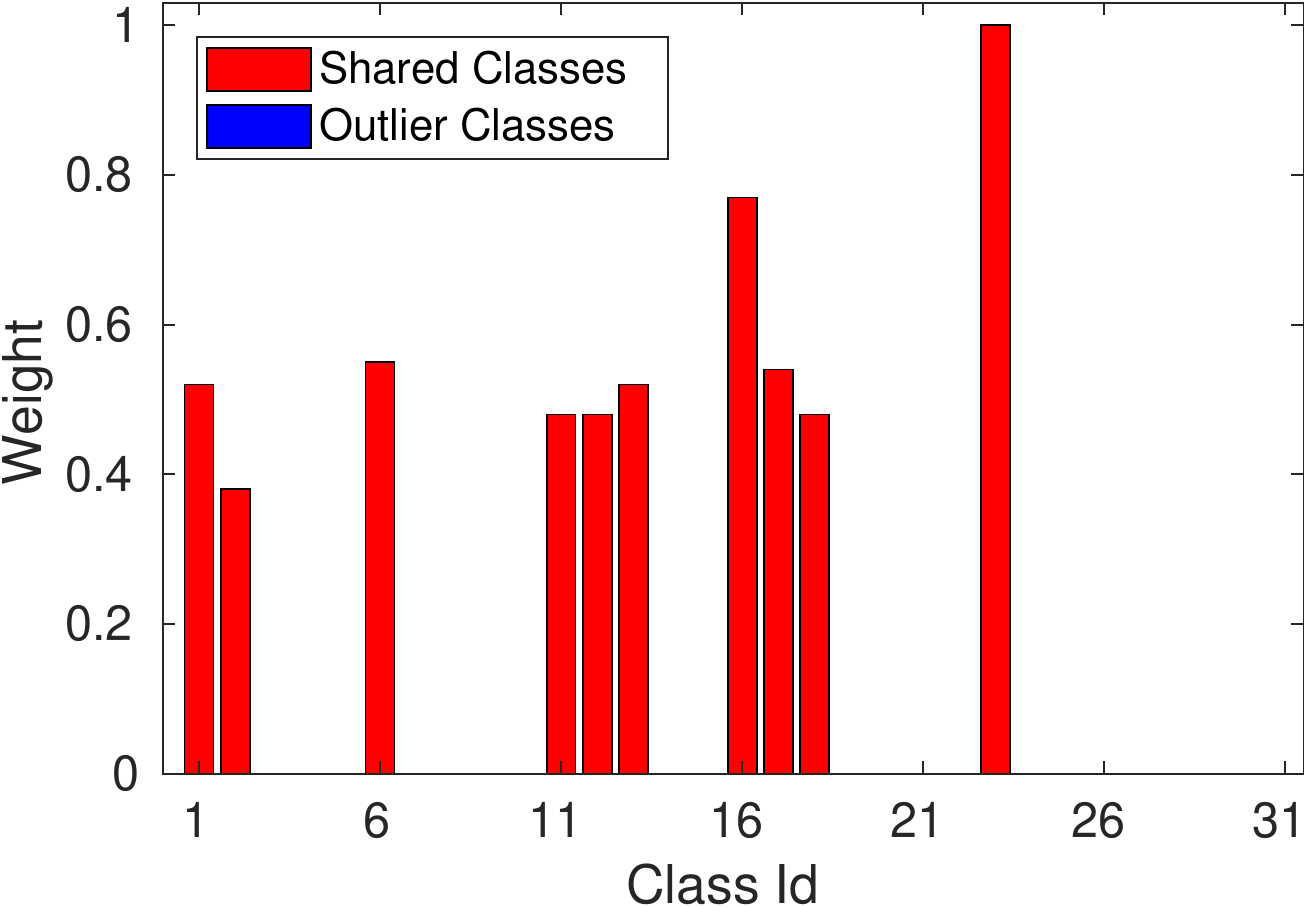}\\
     ~~~PADA  & ~~~SSPDA-$\gamma$ & ~~~SSPDA-PADA\\
\end{tabular}
\caption{Histogram showing the elements of the $\gamma$ vector, corresponding to the class weight learned by PADA, SSPDA-$\gamma$ and SSPDA-PADA for the A$\rightarrow$W experiment.}
\label{fig:bars}
\end{figure}

\subsection{Results of SSPDA combined with  other PDA strategies}
To analyze the combination of SSPDA with the standard PDA source re-weighting technique and the adversarial domain classifier, we extended the experiments on the Office-31 dataset. The bottom part of Table \ref{tab:office31} reports
the obtained results when we add the estimate of the target class statistics through the weight $\gamma$ (SSPDA-$\gamma$) and when also the domain classifier is included in the network as in \cite{PADA_eccv18} (SSPDA-PADA).
In the first case, estimating the target statistics helps the network to focus only on the shared categories, with an average accuracy improvement of two percentage points over the plain SSPDA. Moreover, since the technique to evaluate $\gamma$ is the same used in \cite{PADA_eccv18}, we can state that the advantage comes from a better alignment of the domain features, thus from the introduction of the self-supervised jigsaw task. Indeed, by comparing the $\gamma$ values on the A$\rightarrow$W domain shift we observe that SSPDA-$\gamma$ is more precise in identifying the missing classes of the target (see Figure \ref{fig:bars}). 
In the second case, since the produced features are already well aligned across domains, we fixed $\lambda$-max to $0.1$ and observed a further small average improvement, with the largest advantage when the A domain is used as source. From the last bar plot on the right of Figure \ref{fig:bars} we also observe a further improvement in the identification of the missing target classes.

\section{Conclusions}
In this paper we discussed how the self-supervised jigsaw puzzle task can be used for domain adaptation in the challenging partial setting with some of the source classes missing in the target. Since the high-level knowledge captured by the spatial co-location of patches is unsupervised with respect to the image object content, this task can be applied on the unlabeled target samples and help to close the domain gap without suffering from negative transfer. Moreover we showed that the proposed solution can be seamlessly integrated with other existing partial domain adaptation methods and it contributes to a reliable identification of the categories absent in the target with a consequent further improvement in the recognition results.  
In the future we plan to further explore the jigsaw puzzle task also in the open-set scenario where the target contains new unknown classes with respect to the source. 

%
%
%
\bibliographystyle{splncs04}
\bibliography{egbib}

\begin{thebibliography}{10}
\providecommand{\url}[1]{\texttt{#1}}
\providecommand{\urlprefix}{URL }
\providecommand{\doi}[1]{https://doi.org/#1}

\bibitem{LOAD_ICRA}
Angeletti, G., Caputo, B., Tommasi, T.: Adaptive deep learning through visual
  domain localization. In: ICRA (2018)

\bibitem{Bousmalis:Google:CVPR17}
Bousmalis, K., Silberman, N., Dohan, D., Erhan, D., Krishnan, D.: Unsupervised
  pixel-level domain adaptation with gans. In: CVPR (2017)

\bibitem{SAN}
Cao, Z., Long, M., Wang, J., Jordan, M.I.: Partial transfer learning with
  selective adversarial networks. In: CVPR (2018)

\bibitem{PADA_eccv18}
Cao, Z., Ma, L., Long, M., Wang, J.: Partial adversarial domain adaptation. In:
  ECCV (2018)

\bibitem{jigen}
Carlucci, F.M., D'Innocente, A., Bucci, S., Caputo, B., Tommasi, T.: Domain
  generalization by solving jigsaw puzzles. In: CVPR (2019)

\bibitem{carlucci2017auto}
Carlucci, F.M., Porzi, L., Caputo, B., Ricci, E., Rota~Bul{\`o}, S.: Autodial:
  Automatic domain alignment layers. In: ICCV (2017)

\bibitem{Caruana:1997}
Caruana, R.: Multitask learning. Mach. Learn.  \textbf{28}(1),  41--75 (1997)

\bibitem{deepJDOT}
Damodaran, B.B., Kellenberger, B., Flamary, R., Tuia, D., Courty, N.:
  {DeepJDOT: Deep Joint Distribution Optimal Transport for Unsupervised Domain
  Adaptation}. In: ECCV (2018)

\bibitem{DoerschGE15}
Doersch, C., Gupta, A., Efros, A.A.: Unsupervised visual representation
  learning by context prediction. In: ICCV (2015)

\bibitem{Ganin:DANN:JMLR16}
Ganin, Y., Ustinova, E., Ajakan, H., Germain, P., Larochelle, H., Laviolette,
  F., Marchand, M., Lempitsky, V.: Domain-adversarial training of neural
  networks. J. Mach. Learn. Res.  \textbf{17}(1),  2096--2030 (2016)

\bibitem{Goodfellow:GAN:NIPS2014}
Goodfellow, I., Pouget-Abadie, J., Mirza, M., Xu, B., Warde-Farley, D., Ozair,
  S., Courville, A., Bengio, Y.: Generative adversarial nets. In: NIPS (2014)

\bibitem{slicedWasserstein_cvpr19}
Lee, C.Y., Batra, T., Baig, M.H., Ulbricht, D.: Sliced wasserstein discrepancy
  for unsupervised domain adaptation. In: CVPR (2019)

\bibitem{AI}
Legg, S., Hutter, M.: A collection of definitions of intelligence. Preprint
  arXiv:0706.3639  (2007)

\bibitem{Long_icml15}
Long, M., Cao, Y., Wang, J., Jordan, M.I.: Learning transferable features with
  deep adaptation networks. In: ICML (2015)

\bibitem{Long:2015}
Long, M., Cao, Y., Wang, J., Jordan, M.I.: Learning transferable features with
  deep adaptation networks. In: ICML (2015)

\bibitem{long2016unsupervised}
Long, M., Zhu, H., Wang, J., Jordan, M.I.: Unsupervised domain adaptation with
  residual transfer networks. In: NIPS. pp. 136--144 (2016)

\bibitem{LongZ0J17}
Long, M., Zhu, H., Wang, J., Jordan, M.I.: Deep transfer learning with joint
  adaptation networks. In: ICML (2017)

\bibitem{luo2017label}
Luo, Z., Zou, Y., Hoffman, J., Fei-Fei, L.F.: Label efficient learning of
  transferable representations acrosss domains and tasks. In: NIPS. pp.
  165--177 (2017)

\bibitem{mancini2018boosting}
Mancini, M., Porzi, L., Rota~Bul\`o, S., Caputo, B., Ricci, E.: Boosting domain
  adaptation by discovering latent domains. In: CVPR (2018)

\bibitem{TWIN_PDA}
Matsuura, T., Saito, K., Harada, T.: Twins: Two weighted inconsistency-reduced
  networks for partial domain adaptation. Preprint arXiv:1812.07405  (2018)

\bibitem{morerio2018minimalentropy}
Morerio, P., Cavazza, J., Murino, V.: Minimal-entropy correlation alignment for
  unsupervised deep domain adaptation. In: ICLR (2018)

\bibitem{NorooziF16}
Noroozi, M., Favaro, P.: Unsupervised learning of visual representations by
  solving jigsaw puzzles. In: ECCV (2016)

\bibitem{learningtocount}
Noroozi, M., Pirsiavash, H., Favaro, P.: Representation learning by learning to
  count. In: ICCV (2017)

\bibitem{russo17sbadagan}
Russo, P., Carlucci, F.M., Tommasi, T., Caputo, B.: From source to target and
  back: symmetric bi-directional adaptive gan. In: CVPR (2018)

\bibitem{Saenko:2010}
Saenko, K., Kulis, B., Fritz, M., Darrell, T.: Adapting visual category models
  to new domains. In: ECCV (2010)

\bibitem{sankaranarayanan2017generate}
Sankaranarayanan, S., Balaji, Y., Castillo, C.D., Chellappa, R.: Generate to
  adapt: Aligning domains using generative adversarial networks. In: CVPR
  (2018)

\bibitem{Hoffman:Adda:CVPR17}
Tzeng, E., Hoffman, J., Darrell, T., Saenko, K.: Adversarial discriminative
  domain adaptation. In: CVPR (2017)

\bibitem{Tzeng:MMD:arxiv14}
Tzeng, E., Hoffman, J., Zhang, N., Saenko, K., Darrell, T.: Deep domain
  confusion: Maximizing for domain invariance. Preprint arXiv:1412.3474  (2014)

\bibitem{venkateswara2017Deep}
Venkateswara, H., Eusebio, J., Chakraborty, S., Panchanathan, S.: Deep hashing
  network for unsupervised domain adaptation. In: CVPR (2017)

\bibitem{featurenorm_PDA}
Xu, R., Li, G., Yang, J., Lin, L.: Unsupervised domain adaptation: An adaptive
  feature norm approach. Preprint arXiv:1811.07456  (2018)

\bibitem{Zellinger_CMD17}
Zellinger, W., Grubinger, T., Lughofer, E., Natschl{\"{a}}ger, T.,
  Saminger{-}Platz, S.: Central moment discrepancy {(CMD)} for domain-invariant
  representation learning. In: ICLR (2017)

\bibitem{IWAN}
Zhang, J., Ding, Z., Li, W., Ogunbona, P.: Importance weighted adversarial nets
  for partial domain adaptation. In: CVPR (2018)

\bibitem{zhang2016colorful}
Zhang, R., Isola, P., Efros, A.A.: Colorful image colorization. In: ECCV (2016)

\end{thebibliography}

\end{document}